\documentclass[lettersize,journal]{IEEEtran}
\usepackage{amsmath,amsfonts, physics}
\usepackage{algorithmic}
\usepackage{array}
\usepackage[caption=false,font=normalsize,labelfont=sf,textfont=sf]{subfig}
\usepackage{textcomp}
\usepackage{stfloats}
\usepackage{url}
\usepackage{verbatim}
\usepackage{graphicx}
\usepackage{svg}
\usepackage{hyperref}
\usepackage{enumitem}
\usepackage{multirow}
\usepackage{amssymb}
\usepackage{float}
\newlist{steps}{enumerate}{1}
\setlist[steps, 1]{label = Step \arabic*:}
\graphicspath{ {./images/} }

\def\BibTeX{{\rm B\kern-.05em{\sc i\kern-.025em b}\kern-.08em
    T\kern-.1667em\lower.7ex\hbox{E}\kern-.125emX}}
\usepackage{balance}
\begin{document}
\title{Deep Learning Guided Autonomous Surgery: Guiding Small Needles into Sub-Millimeter Scale Blood Vessels
}
\author{Ji Woong Kim, Peiyao Zhang, Peter Gehlbach, Iulian Iordachita, Marin Kobilarov
\thanks{Ji Woong Kim, Peiyao Zhang, Iulian Iordachita, and Marin Kobilarov are with the Department of Mechanical Engineering and Laboratory for Computational Sensing and Robotics (LCSR), Johns Hopkins University, Baltimore, MD 21218 USA (e-mail: {jkim447, pzhang24, iordachita, marin}@jhu.edu).

 Peter Gehlbach is with the Wilmer Eye Institute, Johns Hopkins University Hospital,
Baltimore, MD 21287 USA (e-mail: pgelbach@jhmi.edu)}}

\markboth{Journal of \LaTeX\ Class Files,~Vol.~18, No.~9, September~2020}%
{How to Use the IEEEtran \LaTeX \ Templates}

\maketitle
    


\begin{abstract}
We propose a general strategy for autonomous guidance and insertion of a needle into a retinal blood vessel. The main challenges underpinning this task are the accurate placement of the needle-tip on the target vein and a careful needle insertion maneuver to avoid double-puncturing the vein, while dealing with challenging kinematic constraints and depth-estimation uncertainty. Following how surgeons perform this task purely based on visual feedback, we develop a system which relies solely on \emph{monocular} visual cues by combining data-driven kinematic and contact estimation, visual-servoing, and model-based optimal control. By relying on both known kinematic models, as well as deep-learning based perception modules, the system can localize the surgical needle tip and detect needle-tissue interactions and venipuncture events. The outputs from these perception modules are then combined with a motion planning framework that uses visual-servoing and optimal control to cannulate the target vein, while respecting kinematic constraints that consider the safety of the procedure. We demonstrate that we can reliably and consistently perform needle insertion in the domain of retinal surgery, specifically in performing retinal vein cannulation. Using cadaveric pig eyes, we demonstrate that our system can navigate to target veins within 22$\mu m$ XY accuracy and perform the entire procedure in less than 35 seconds on average, and all 24 trials performed on 4 pig eyes were successful. Preliminary comparison study against a human operator show that our system is consistently more accurate and safer, especially during safety-critical needle-tissue interactions. The principles of the proposed system may also be applicable to other surgical domains requiring surgical tool guidance and venipuncture, or in general, vision-based robotic manipulation tasks. To the best of the authors' knowledge, this work accomplishes a first demonstration of autonomous retinal vein cannulation at a clinically-relevant setting using animal tissues.
\end{abstract}


\begin{figure*}[h]
        \centering
        \includegraphics[width = \textwidth]{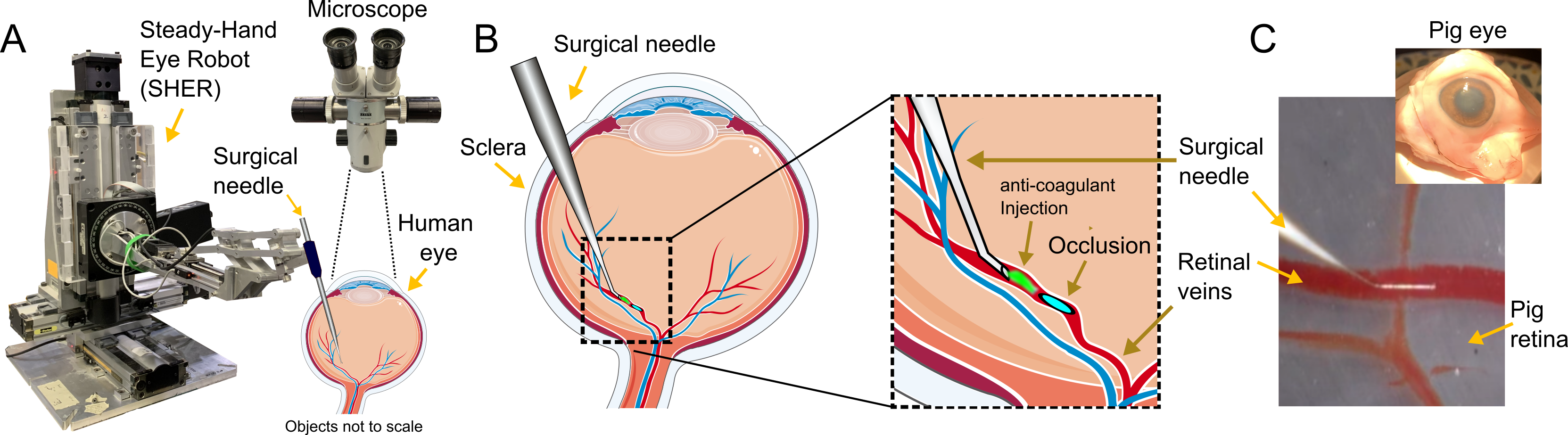}
        \caption{\small (A) Illustration of retinal surgery; a surgical needle is inserted into the eye via a point incision on the sclera. A microscope is suspended over the patient's eye to visualize the retina. (B) Illustration of retinal vein cannulation procedure; a microneedle is penetrated into a small blood vessel (typically ranging from 60 - 120 $\mu$m) to inject anti-coagulants and dissolve an occlusion downstream of the vein. (C) Cadaveric pig eye used in the experiments}
        \label{fig:intro_pic}
\end{figure*}

\section{INTRODUCTION}

Retinal vein occlusion (RVO) is the second-most common retinal vascular disorder, affecting $\sim$16.4 million adults world-wide \cite{rvo_prevalence}. RVO is caused by a blockage in the retinal veins or venules, restricting blood flow and oxygen delivery to the retina. Physiological consequences include retinal ischemia, that in turn leads to death of severely affected tissue, or in surviving tissue compromised of vascular integrity. Retinal edema, hemorrhage, and neovascularization are notable consequences of chronic retinal iscehmia. Loss of vision, retinal detachment, glaucoma, and loss of the eye are known clinical outcomes.

Current treatments for RVO focus on mitigating its symptoms. Common treatments include intravitreous injection of anti-VEGF agents to limit vascular leakage and prevent neovascularization. However, these costly treatments are often required monthly, may not preserve vision, and do not address the root cause of the problem. A potential definitive treatment for RVO is retinal vein cannulation (RVC). RVC is a surgical procedure during which a microneedle is inserted into the vein, upstream to the blockage, and clot dissolving anti-coagulants are injected (Fig. \ref{fig:intro_pic}). The injection chemically dissolves the blockage, removing the physical obstruction inside the vein. The result is restoration of blood flow to the retina. However, due to the micron-scale and fragility of the venules, RVC is technically difficult to perform safely. It requires complex surgical skills exceeding the physiological limits of many surgeons. Therefore, RVC has not become a standard of care.

\setcounter{figure}{1}
\begin{figure*}[h]
        \centering
        \includegraphics[width = \textwidth]{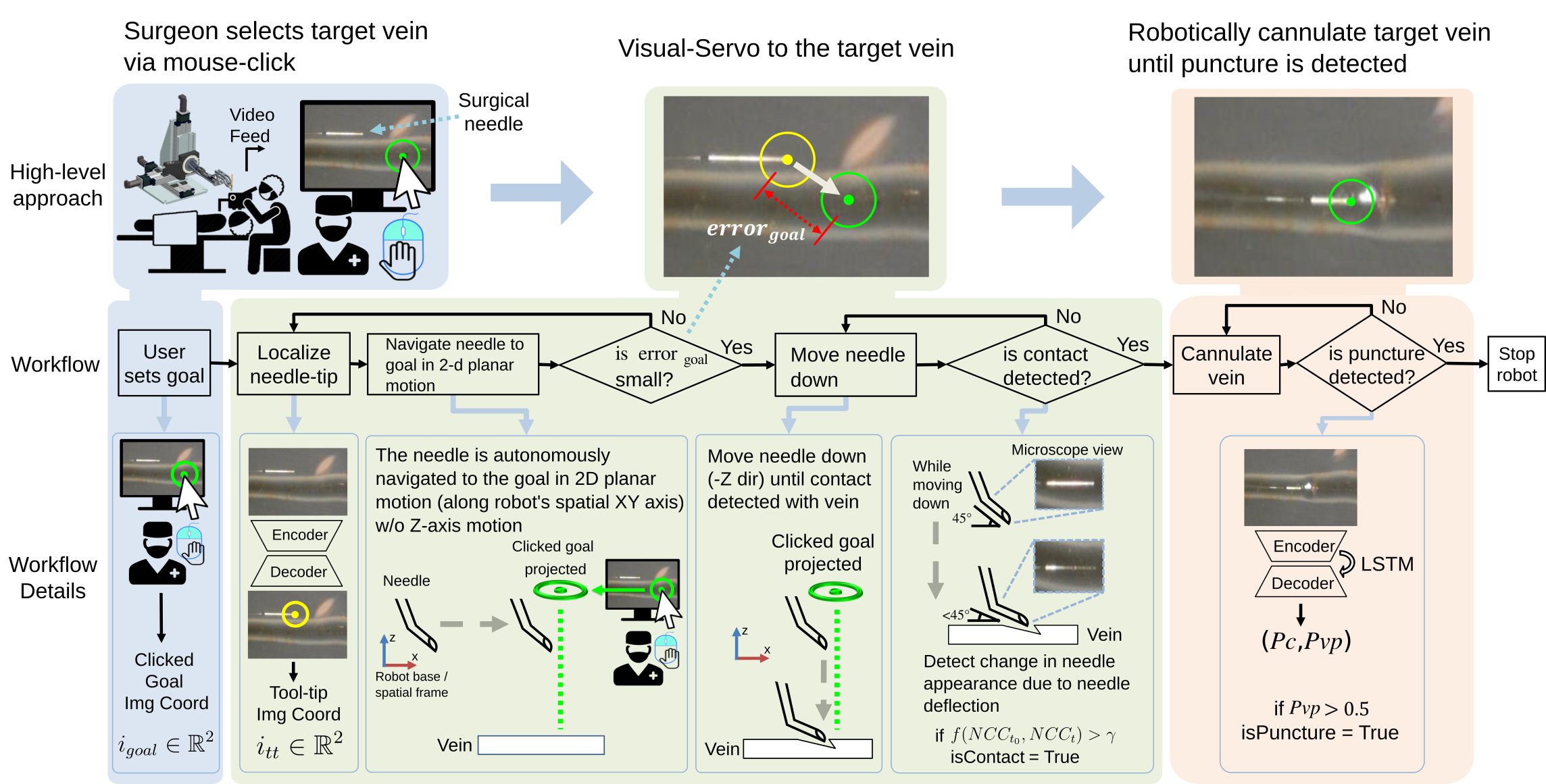}
        \caption{\small Illustration of the overall workflow: the surgeon specifies the needle insertion point by directly clicking on the visual feed of the surgery. Then, the needle is autonomously navigated to the goal via a specific strategy where the motion first consists of 2-d planar motion without moving down, followed by a moving-down motion to reach the goal, which is an efficient way to reach any point on the retinal tissue. The needle is moved down until a contact event between the needle-tip and the target vein is detected via a visual cue. The needle is then inserted into the vein until a puncture event is detected via a trained CNN. All procedures beyond the initial goal-selection step is performed autonomously and no further user intervention is necessary.}
        \label{fig:workflow_part_a}
\end{figure*}

During RVC, the surgeon uses a binocular stereo surgical microscope to visualize the retina directly through the cornea (Fig. 1). Surgical tools are inserted through scleral ports and the workspace is illuminated using a light pipe. During surgery, extreme precision is required. Targeted blood-vessels range from  60 - 120$\mu$m  in diameter  (approximately the size of the human hair diameter), and the required accuracy for needle placement is on the order of 20 - 30$\mu$m. It is extremely difficult to guide sharp surgical instruments with such accuracy, while ensuring that the targeted tissue is not damaged during cannulation. Furthermore, the observed mean amplitude of human free-hand tremor during retinal surgery is on the order of 200$\mu$m \cite{hand_tremor_amplitude}. Consistently achieving the required precision using free-hand guidance is therefore extremely challenging, even for experienced surgeons. Depth guidance is a further challenge during needle insertion. The surgeon must be careful not to double-puncture through both walls of the vein. A double-puncture may cause a hemorrhage, extravasation of drug on injection, or traumatically damage the underlying non-regenerative tissue of the retina. Additionally, because the surgery is performed from a top-down view using a microscope, depth perception is severely limited. Such uncertainty increases the risk for error during needle placement and cannulation.

\begin{figure*}[h]
        \centering
        \includegraphics[width = 0.87\textwidth]{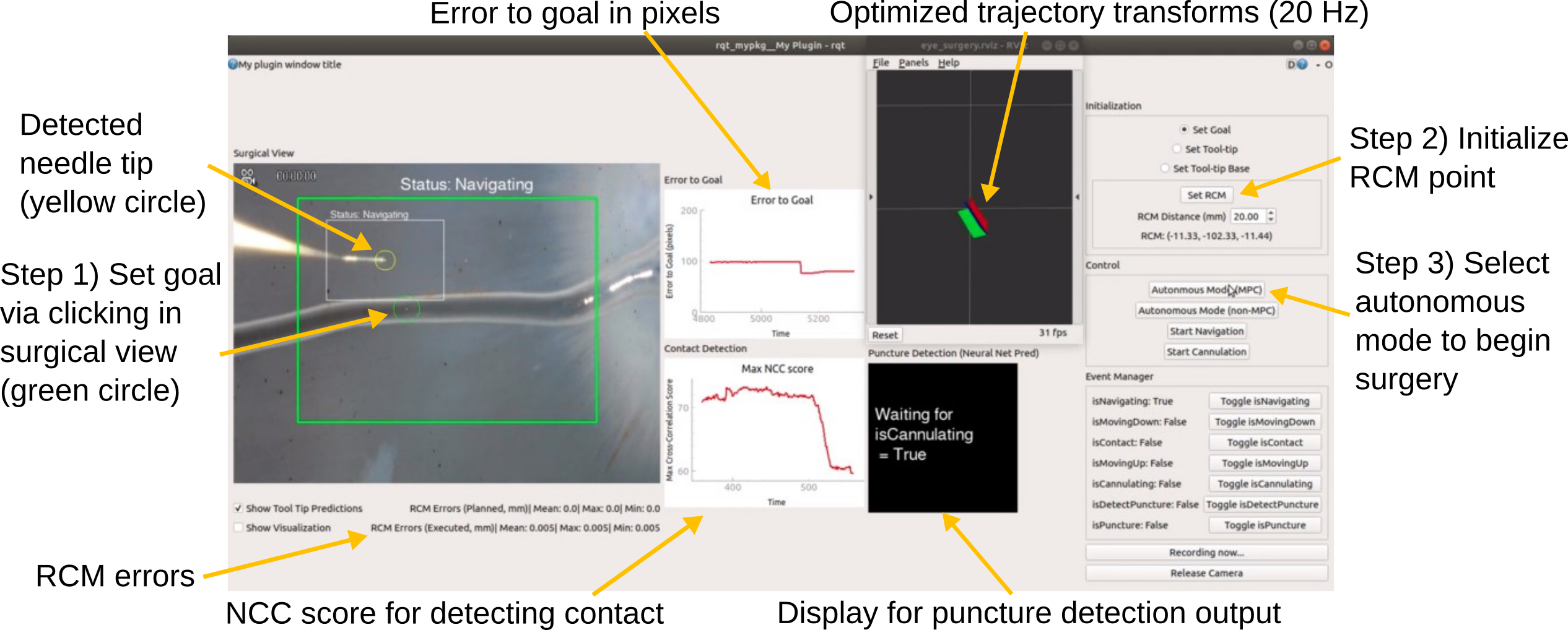}
        \caption{\small  GUI for interacting with the system and monitoring the progress of the surgery; only three mouse clicks are necessary to initalize the system: 1) goal selection on the target vein 2) RCM point initialization 3) start button}
        \label{fig:gui}
\end{figure*}

Several recent technological innovations have been introduced to alleviate the difficulty of retinal surgery. Notably, robotic assistance has been shown to reduce the surgeon's hand tremor and enhance the accuracy and precision during surgical tool navigation (see Section \ref{related_works}). However, robotic assistance does not yet free the surgeon from having to make complex split-second decisions during surgery. It also does not address the uncertainty of depth estimation during surgery. Furthermore, the surgeon needs to be trained to become an expert in using each robotic platform, which requires time and resources. Thus, an automated solution to RVC could help surgeons focus on less-technically demanding aspects of the surgery. Eventually, automated surgical care may increase access to care at a lower cost with improved surgical outcomes.

In this paper, we propose a first automated solution to RVC. Specifically, we develop a system directed at automating the following tasks: (1) navigating the surgical needle to the target vein, (2) inserting the needle into the vein, and (3) detecting a venipuncture event to avoid double-puncturing the vein. We accomplish these tasks using minimal setup and guidance from the surgeon. Specifically, our system only requires two inputs: top-down monocular images from the microscope for visual feedback and a 2-d pixel goal indicating the needle insertion point. This needle insertion point is specified by the surgeon via a simple mouse-click in the visual feed of the surgery, as illlustrated in Fig. \ref{fig:workflow_part_a}. No other input or human intervention is necessary. Notably, there is no need to estimate depth using depth-measuring sensors or explicitly calculate the relative transform between the needle tip and the target tissue. Only visual cues obtained from monocular images are used to automate RVC. 

At a high-level, we solve the needle navigation problem by simply guiding the needle towards the target vein until a needle-vein contact cue can be \emph{visually} observed. This is accomplished by performing navigation to the clicked goal via 2-d visual-servoing initially (i.e. moving only along the robot's spatial XY plane), followed by a lowering motion towards the target vein (i.e. moving only along the robot's spatial Z-axis), as illustrated in Fig. \ref{fig:workflow_part_a}. As the needle approaches the target vein, a vision module tracks the needle to detect a needle-vein contact event, so that the robot can be stopped upon detection. Specifically, when the needle tip contacts the target vein, the needle tip deflects and changes appearance. This specific visual variation is used as a contact cue to stop the robot. This approach is analogous to how surgeons rely on visual cues to detect contact events during retinal surgery. Once the needle is gently placed on the target vein, the needle is actuated along its axis to pierce the vein. An important consideration during this step is avoiding a double-puncture of the vein. Therefore, during needle insertion, a recurrent convolutional neural network (CNN) is used to detect puncture events from a sequence of images in real-time. Once puncture is detected by the recurrent CNN, the robot is automatically stopped and held in place for sustained injection of therapeutic anticoagulant drug. Throughout the entire procedure, an optimal control framework is used in tandem with visual-servoing pipeline to generate smooth trajectories while respecting the remote center of motion (RCM) constraint at the scleral entry point. Specifically, the visual-servoing pipeline generates a goal waypoint to reach, which is used in the optimal control formulation to generate an optimized trajectory to the goal based on specified costs and constraints. Notably, the optimized trajectory ensures that the tool's axis is always aligned with the entry point of the tool, thereby avoiding extraneous rotations on the eye.

We note that the above strategy borrows from how surgeons perform RVC. It is known that surgeons rely primarily on visual feedback to perform retinal surgery, even when interacting with the retinal tissue. This is the direct result of the fact that most tool-to-tissue interaction forces are below the level of human tactile perception. Notably, events such as tool-tissue contact and venipuncture are detected visually rather than via tactile feedback (e.g. by visually observing the deformation of the tissue). Additionally, an interesting observation is that some surgeons do not even use stereo microscopes during retinal surgery, which is known to help with depth estimation via stereopsis. Instead, some surgeons increasingly utilize \emph{monocular} heads-up displays to perform retinal surgeries. In other words, monocular images already contain much of the information necessary, and arguably stereo vision is not necessary for depth estimation in all retinal surgery situations. This observation motivated us to develop an autonomous RVC system that relies only on monocular image input, and in particular relying on visual cues to detect tool-tissue interactions during vein contact and puncture. This simplification eliminates the need to estimate depth explicitly, by using alternate approaches such as stereo vision or expensive sensors like optical coherence tomography (OCT).

Using the proposed strategy, we demonstrate the feasibility of our autonomous RVC system using "open-sky" cadaveric pig eyes (i.e. the anterior segment structures have been removed for greater accessibility). Pig eyes are a popular choice of animal model preceding clinical trials due to their similarity in size and tissue anatomy to the human eye. Since the eye is detached from the heart and lacks pressure inside the veins, we injected air into the veins to simulate intraluminal venous pressure, as shown in Fig. \ref{fig:pig_eye}. To test the consistency of our system, we performed a total of 24 needle insertions on 4 different pig eyes. The cannulations were performed in back-to-back manner to test its unbiased accuracy and consistency. We report that all 24 trials were successful based on the following criteria: the needle was navigated to the target with very little error, puncture was successfully observed, and the puncture was detected in a timely manner to stop the robot immediately without exceeding the margin of safety.

\section{Related Works} \label{related_works}

Over the past two decades, numerous technologies have been introduced to assist retinal surgeons. Various robot-assisted platforms have been introduced, including hand-held devices, tele-operated systems, and cooperative systems \cite{robotic_surgery_review}. Robotic assistance has been shown to be effective in reducing a surgeon's hand-tremor and to improving accuracy in navigating surgical tools. Recent notable robotic platforms were developed by Mynutia and Preceyes, which have reached clinical trials \cite{first_human_retinal_surgery},  \cite{first_human_subretinal_injection}. The Mynutia system has been used to perform RVC on humans, while the surgeon was in control of the robot throughout the entire procedure. These assistive platforms can be further augmented with sensors such as force-sensing at the tool-tip using Fiber Bragg Gratings (FBG) sensors \cite{fbg_force_sensing}, depth-sensing using OCT-integrated surgical tools \cite{OCT_integrated_tool} \cite{OCT_integrated_tool_2}, and impedance-measuring sensors for detecting venipuncture events \cite{puncture_detection_impedance}. Existing literature provides a thorough review of recent advancements of robot-assisted systems and sensors used in retinal surgery \cite{robotic_surgery_review}. In the present paper, we focus our discussion on efforts related to automation.

Recent works related to automation aim to regress relevant information required for automation (e.g. depth estimation) or solve a sub-task within a multi-step procedure (e.g. tool navigation). Depth estimation is relevant for finding the relative transform between the needle-tip and the target tissue, which can then be utilized for navigating surgical tools. For example, \cite{stereo_proximity_detection} used stereo vision to estimate the depth of the tool and the retina individually to develop a proximity detection system. In similar work by \cite{forcep_landmark_detection}, keypoints were detected from surgical tools to estimate camera intrinsic and extrininsic properties, which were then used to perform a rough reconstruction of the retinal surface. However, current stereo vision methods simply do not provide depth reconstruction accuracy sufficient for automation. An alternative direction for automation relies on using tool-shadow dynamics as cues to estimate depth. Specifically, when the surgical tool is placed near the retina, the physical tool-tip and its shadow-tip converge, which can be used as cues to approximate proximity. For example, \cite{japan_shadow} used tool-shadow cues to warn surgeons of proximity when the tool-tip and the shadow-tip converged by a predefined pixel distance. An extension of this work explored using deep networks to imitate surgical actions performed by an expert, while using tool-shadow dynamics as visual cues for depth. Specifically, \cite{autonomous_navigation_retina}, \cite{eye_surgery_imitation_learning}, and \cite{peiyao_eye_surgery} combined deep imitation learning with optimal control to generate trajectories that imitated expert actions based on the observed tool-shadow dynamics in the microscope image. These methods were shown to be able to navigate surgical tools to desired locations on the retina with $\sim$100 $\mu$m accuracy on silicone eye phantoms and in a graphics simulator using Unity3D. \cite{peiyao_pig_eye_paper} further proved its efficacy on pig closed eyes, while ensuring greater safety using chance constraints. However, the navigation accuracy along the depth dimension was still insufficient for very precise needle placements, which is required for vein cannulation application.


Intraoperative optical coherence tomography (iOCT) recently emerged as a promising candidate for enabling surgical automation. iOCT is a laser-scanning imaging modality that provides accurate reconstruction of tissue geometry with several micron accuracy. Several works have demonstrated promising results with iOCT in retinal surgery. Notably, \cite{autonomous_rvc_silicone} demonstrated OCT-guided vein cannulation using a silicone phantom. Others have demonstrated the feasibility of using iOCT for micro-suturing on a silicone phantom \cite{oct_micro_suture} and performing corneal kerotoplasty on \emph{ex-vivo} human cornea \cite{oct_keratoplasty}. While iOCT provides depth measurements that are sufficiently accurate for autonomous applications, a major drawback is its slow image acquisition and refresh rates due to the large amount of data processing involved. Thus, the aforementioned works were limited to using only cross-sectional scans (i.e. B-scans) rather than full 3-d volume tissue scans to circumvent issues with slow image acquirement. This workaround is problematic as it limits the workspace to a single dimension (i.e. along the B-scan plane) and the slightest motion of the patient can displace the target or the tool from the imaging plane. Another limitation of iOCT is its high cost and hence the lack of accessibility in many clinics. In this work, we avoid the assumption that depth estimation is required and demonstrate that only visual cues are necessary to accomplish automation for RVC. Therefore, we have developed our system using only monocular image input, and do not require stereo vision or OCT sensors to acquire depth measurements. Furthermore, microscope cameras are widely available in clinical practices and its images can be obtained at high frame rates. This enables us to develop a system that can respond to real-time changes in the surgical environment and potentially roll-out our system into clinical application.   

\begin{figure*}[h]
        \centering
        \includegraphics[width = 0.84 \textwidth]{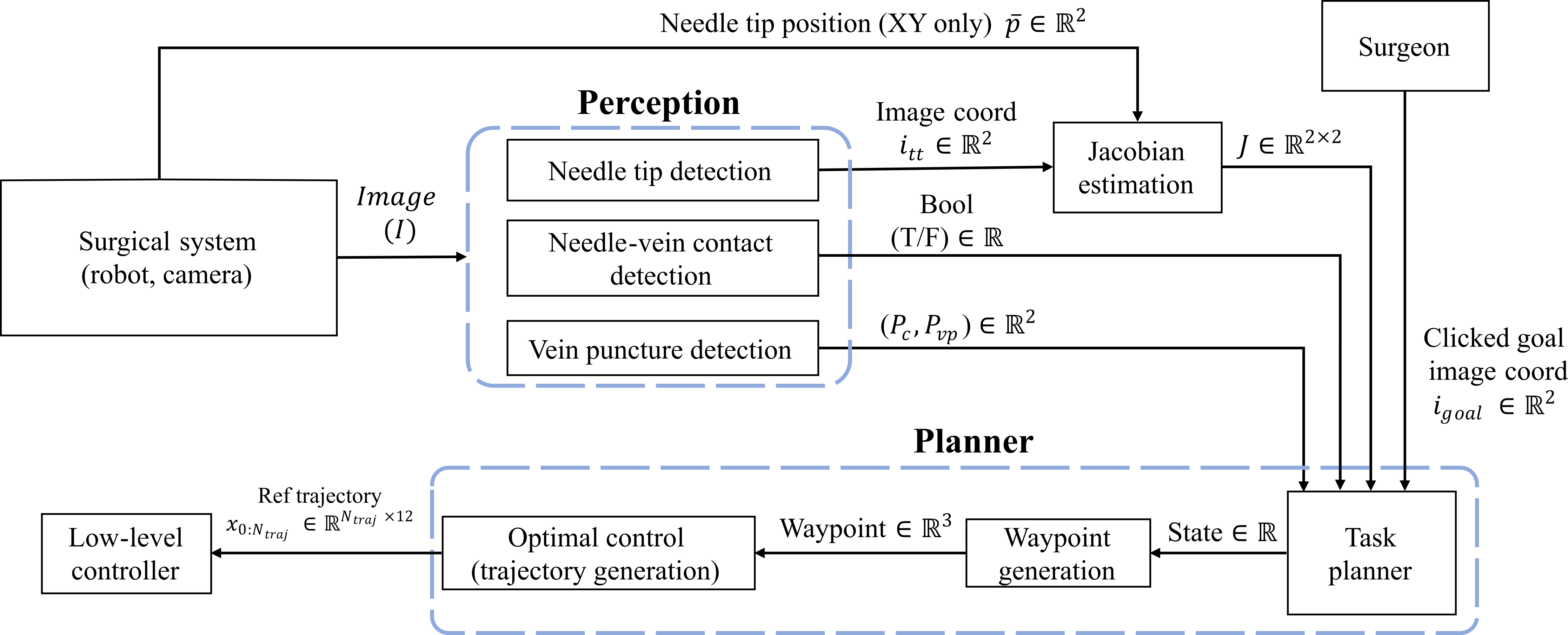}
        \caption{\small Data-centric illustration of the system; the system largely consists of the perception module and the planner module.}
        \label{fig:workflow_prt_b}
\end{figure*}

\begin{figure}[H]
        \centering
        \includegraphics[width = \columnwidth]{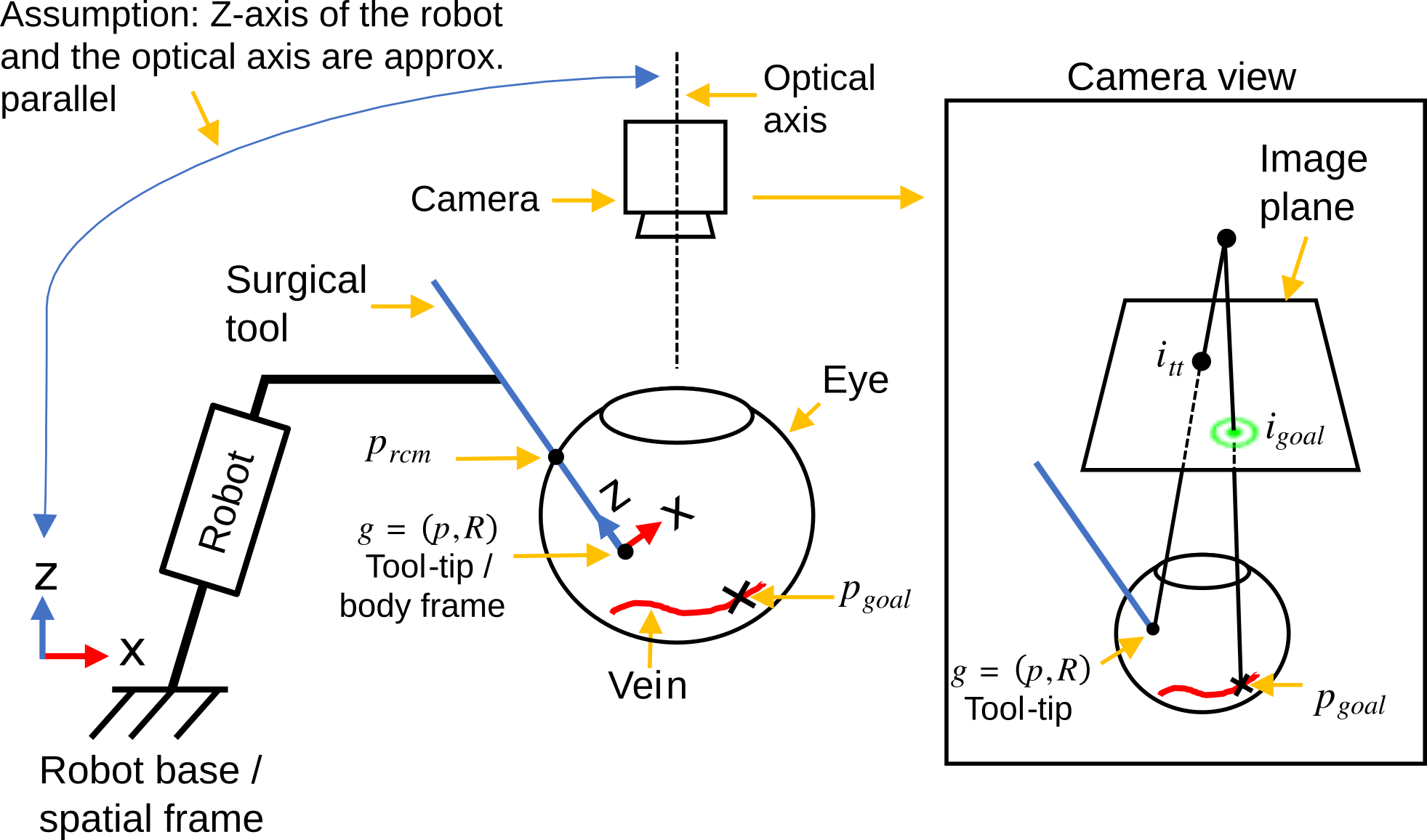}
        \caption{\small  Visual illustration of the variables used}
        \label{fig:surgical_setup_variables}
\end{figure}

\section{Problem Formulation} \label{problem_formulation}
Consider a robotic manipulator with an end-effector (i.e. surgical tool-tip) states defined as $x=(g, V)$, where $g = (p, R) \in SE(3)$ and $V = (v, \omega) \in \mathbb{R}^6$. $p$ denotes the tool-tip position, $R\in SO(3)$ the orientation, and $v\in\mathbb{R}^3$ and $\omega\in\mathbb{R}^3$ the end-effector-frame translational and angular velocity. Let the robot end-effector occupy a region $A(q)\subset \mathcal W$ in the workspace $\mathcal W\subset \mathbb{R}^3$. The key variables to be discussed are visually illustrated in Fig. \ref{fig:surgical_setup_variables}.

The end-effector state $x$ is fully-observable using high-precision motor encoders and precise knowledge of the robot forward kinematics. Forces applied at the end-effector (or optionally desired end-effector velocity) can be mapped to robot joint velocity to move the robot. Additionally, the system includes a monocular microscope camera  generating top-down observations of the surgical environment $o(t)\in \mathcal I$ from space of images $\mathcal I$. We assume that the intrinsic and extrinsic properties of the camera are unknown.

Initially, the surgeon manually introduces the surgical tool into the eye through a sclera entry point, $p_{rcm}\in\mathbb{R}^3$, which is recorded at the time of entry. The sclera point should remain fixed after each entry during the surgery to avoid unsafe forces exerted on the sclera tissue. The surgical tool-tip is observed in the microscope image and is denoted as $i_{tt} \in \mathbb{R}^2$.  The surgeon then selects a 2-d pixel goal $i_{goal} \in o(t)$ on the target vein via a mouse-click in the monocular visual feed of the surgery. This clicked point corresponds to the 3-d euclidean point, $p_{goal} \in \mathcal{W}$, on the surface of the target vein w.r.t the robot's spatial frame. However, $p_{goal}$  is unknown since the surgeon only specifies a 2-d pixel goal from the camera view.

The objective is to navigate the surgical needle near the target vein, gently land the needle on the target vein, perform needle insertion, and during the insertion, detect a venipuncture event to avoid double-puncturing the vein. We thus consider the following three problems:

\begin{enumerate}
\item \emph{Navigating the needle to the target vein}: navigate the surgical tool-tip to the specified needle insertion point, $p_{goal}$, given only 2-d pixel goal, $i_{goal}$.

\item \emph{Detecting needle-vein contact}: detect the moment at which the needle contacts the vein and stop the robot. At this point, the needle should be gently placed on the target vein (as close as possible to $p_{goal}$), ready to be inserted.

\item \emph{Needle insertion and puncture detection}: perform needle insertion and during the insertion, detect a venipuncture event to stop the robot and avoid double-puncturing the vein.
\end{enumerate}

The objective is to autonomously perform the above three tasks while relying solely on monocular images for visual feedback. Additionally, kinematic constraints concerned with the safety of the surgery must be satisfied while ensuring smooth robot motion. We also note that the underlying assumption here is that only visual cues are necessary to solve RVC, without needing to geometrically reconstruct the 3-d state of the environment.

\section{Technical Approach}
The following sections give a technical overview of the RVC system. For organization, we list the relevant sections for the three main tasks mentioned in Section \ref{problem_formulation} in Table \ref{tab:table_sections}. The numerical values for the parameters used are listed in Table \ref{tab:table_parameters}. Fig. \ref{fig:workflow_prt_b} gives a high-level overview of the system from a data-centric point of view.

\begin{table}[hbt!]
\begin{center}
\caption{Main tasks and their relevant sections}
\label{tab:table_sections}
\begin{tabular}{|l|c|}
\hline
\multicolumn{1}{|c|}{Task}                  & Sections \\ \hline
1. Navigating the needle to the target vein & \ref{navigation_logic}, \ref{calibration}, \ref{optimal_control}      \\ \hline
2. Detecting needle-vein contact            & \ref{contact_detection_section}     \\ \hline
3. Needle insertion and puncture detection  & \ref{optimal_control}, \ref{puncture_detection}     \\ \hline
Sections shared across all tasks & \ref{robot_control}, \ref{tool_tip_detection} \\ \hline
\end{tabular}
\end{center}
\end{table}

\begin{table}[hbt!]
\begin{center}
\caption{Parameters used}
\label{tab:table_parameters}
\begin{tabular}{|c|c|c|}
\hline
Parameter & Value   & Description                 \\ \hline
$\alpha$     & 1 pixel & Goal error tolerance (Section \ref{navigation_logic})       \\ \hline
$\beta$      & 0.5     & Jacobian update step-size (Section \ref{calibration})  \\ \hline
$\eta$       & 40$\mu$m & Needle lowering distance (Section \ref{calibration}) \\ \hline
$N_{traj}$ & 64 & Number of trajectory waypoints used (Section \ref{optimal_control}) \\ \hline
$\gamma$     & 0.18    & Contact detection threshold (Section \ref{contact_detection_section}) \\ \hline
\end{tabular}
\end{center}
\end{table}
\subsection{Navigation Logic} \label{navigation_logic}
 
 A fundamental challenge during surgical tool navigation is mapping the user-defined 2-d pixel goal  ($i_{goal}$) to a corresponding 3-d goal point on the surface of the blood vessel ($p_{goal}$), as illustrated in Fig. \ref{fig:surgical_setup_variables}. If $p_{goal}$ can be estimated accurately, then the needle-tip can be simply navigated to the target vein and the navigation problem would be trivially solved. However, in practice, $p_{goal}$ is difficult to estimate accurately. Thus, we circumvent this problem altogether and propose a navigation strategy which the clicked goal can be reached without knowing its 3-d position. The proposed strategy is as follows: 
 
 \begin{enumerate}[wide=0pt, leftmargin=*, label= Step \arabic*)]
     \item Align the needle-tip with the clicked goal-pixel via 2-d visual-servoing i.e. via actuation only along the robot's spatial XY plane as illustrated in Fig. \ref{fig:workflow_part_a}. This step effectively aligns the needle-tip with the clicked goal pixel in the microscope view. 
     
     \item Lower the needle towards the target vein via incremental motion along the robot's spatial Z-axis, as illustrated in Fig. \ref{fig:workflow_part_a}

    \item Repeat the above steps until contact between the needle-tip and the target vein is observed.

\end{enumerate}

An example of a trajectory generated by this navigation procedure is shown in Fig. \ref{fig:comparison_results}D. The proposed strategy assumes that the optical axis of the camera and the robot's spatial Z-axis are approximately parallel (Fig. \ref{fig:surgical_setup_variables}), which is usually the case during retinal surgery. Therefore, during 2-d  planar motion (step 1), the observed motion of the needle-tip is a corresponding planar motion in the image. During Z-axis motion (step 2), the observed motion of the needle-tip is nearly static in the image. This is again due to the Z-axis of the robot being approximately parallel to the optical axis of the camera. However, since the two axes are not perfectly parallel, during Z-axis motions, small lateral motion of the needle-tip may be observed in the image. This results in the needle-tip deviating from the clicked goal. To realign the needle-tip with the clicked goal, small XY adjustments are made whenever the observed deviation is beyond a pixel threshold of $\alpha$. Ultimately, this procedure enables a very accurate placement of the needle-tip on the target blood vessel, without explicitly knowing the target vein location in 3-d. In general, this navigation strategy can be employed to reliably guide the needle-tip anywhere on the retinal surface. 

\subsection{Hand-eye Calibration and Visual Servoing}  \label{calibration}

An important consideration during navigation is ascertaining the calibration matrix between the robot and the camera for visual-servoing (i.e. hand-eye calibration). To accomplish this, we choose an uncalibrated visual servoing strategy which enables real-time adaptation to changes to the intrinsic and extrinsic properties of the camera \cite{jagersand_vs}. This enables the surgeon to change the optical settings (e.g. magnification), reposition the camera, or alter various optical components between the cornea and the microscope (e.g. readjust the BIOME lens or contact lens) during the surgery, without needing to cycle through a dedicated calibration procedure repeatedly. 

Since we assume that the robot's spatial Z-axis and the camera's optical axis are approximately parallel, the calibration is only performed between the robot's spatial XY plane and the image plane. Thus, we define a variable $\bar{p} = Sp \in \mathbb{R}^2$, which denotes the XY components of the surgical tool-tip position, where the selector matrix $S \in \mathbb{R}^{2 \cross 3}$ picks out the first two elements of the vector it operates on. $S$ is defined as:
\begin{align}
S = \begin{bmatrix}
1 & 0 & 0 \\
0 & 1 & 0 
\end{bmatrix}.
\end{align}
The tool-tip is visualized in the camera and its corresponding projection point on the image plane is given as $i_{tt}$, as shown in Fig. \ref{fig:surgical_setup_variables}. Consider an unknown function $K:\mathbb{R}^2 \rightarrow \mathbb{R}^2$ which converts the tool-tip XY position to its corresponding image coordinates. $K$ implicitly contains the intrinsic and extrinsic camera parameters of the camera. In other words,
\begin{equation}
i_{tt} = K(\bar{p}).
\end{equation}
Since $K$ is unknown, we may approximate it using a first-order Taylor series approximation:
\begin{align}
 K(\bar{p}^{k+1}) &\approx K(\bar{p}^k) + J_{calib}(\bar{p}^k)(\bar{p}^{k+1}-\bar{p}^k) \\ 
\Delta i_{tt}^k & \approx J_{calib}(\bar{p}^k)(\Delta \bar{p}^k) \label{eq:jacobian_eq}
\end{align}
where $J_{calib}$ is a jacobian matrix that relates the change in tool-tip XY position, $\Delta \bar{p}^k$, to its change in image coordinates, $\Delta i_{tt}^k$. $k$ denotes the iteration step, since the jacobian is a local approximation at a particular time step. Specifically, the jacobian is recalculated whenever a significant change in both $\Delta i_{tt}^k$ and $\Delta \bar{p}^k$ is observed (i.e. when robot motion observed). This constant recalculation enables real-time adaptation to changes to the aforementioned extrinsic and intrinsic properties of the camera.

One simple way to determine the jacobian matrix is to experimentally perform orthogonal motions on the robot and manually populate its elements using finite differences. However, this introduces unnecessary calibration motions unrelated to the desired task and require re-execution when the intrinsic or extrinsic properties of the camera change. Instead, we use an online method which estimates the jacobian by observing robot motions on the fly, without introducing any unrelated calibration motions \cite{jagersand_vs}. The online method uses Broyden's update formula to estimate the jacobian, which is given as:
\begin{equation} \label{eq:broyden}
    J_{calib}^{k+1} = J_{calib}^{k} + \beta \frac{\Delta i_{tt}^k - J_{calib}^k \Delta \bar{p}^k}{(\Delta \bar{p}^k )^T (\Delta \bar{p}^k)} (\Delta \bar{p}^k)^T,
\end{equation}
where $0 \leq \beta \leq 1$ is the step size for updating the jacobian \cite{Broydens_method}. Note that Broyden's method yields an updated jacobian  that satisfies the secant condition (i.e. $\Delta i_{tt}^k = J_{calib}^{k+1} \Delta \bar{p}^k$) while minimizing the difference between the jacobian from the previous iteration w.r.t the Frobenius norm (i.e. $\norm{J_{calib}^{k+1} - J_{calib}^k}_F$). The initial jacobian may be initialized as an arbitrary non-singular matrix such as an identity matrix or precomputed using prior data for faster convergence. To use the jacobian to guide the needle to the given pixel goal, we reformulate Eq. \ref{eq:jacobian_eq} as
\begin{equation} \label{eq:visual_servoing}
    \Delta \bar{p}_{desired}= J_{calib}(\bar{p}^k)^{-1} \Delta i_{desired},
\end{equation}
where $\Delta i_{desired} = i_{goal} - i_{tt}$ or the desired motion vector in image coordinates, and $\bar{p}_{desired}$ is the desired change in tool-tip position to align the needle-tip with the clicked goal. In case the jacobian is not invertible, an appropriate pseudoinverse may be computed or the most recent non-singular jacobian may be reused from previous iterations. In practice, given the orthogonal nature of the observed motions, it is rare that the jacobian is singular.

Using $\Delta \bar{p}_{desired}$, the desired waypoint to reach can be given as:

\begin{equation} \label{eq:visual_servoing}
    p_{desired}= p + \begin{bmatrix}
           \Delta \bar{p}_{desired} \\
          0 \\ 
           \end{bmatrix}.
\end{equation}
To be clear, note that $p_{desired} \neq p_{goal}$. $p_{desired}$ is an intermediate waypoint that is constantly updated such that the tool-tip will be eventually aligned with the clicked goal in the image view.

Recall that when the needle-tip is aligned with the clicked goal, the next step is to move the robot downwards (along the robot's spatial Z-axis) toward the target vein. During this step, the desired waypoint can be given as:

\begin{equation} \label{eq:needle_lowering}
    p_{desired}= p + \begin{bmatrix}
           0 \\
          0 \\
          -\eta \\
           \end{bmatrix}
\end{equation}
where $\eta$ is simply the distance by which the needle is lowered along the robot's spatial Z-axis.

Depending on whether the required motion is 2-d navigation or needle-lowering, $p_{desired}$ is determined accordingly either using Eq. \ref{eq:visual_servoing} or Eq. \ref{eq:needle_lowering}. Once $p_{desired}$ is determined, it is then used as a goal waypoint in the optimal control framework, which we describe next.

\begin{figure}[h]
        \centering
        \includegraphics[width = \columnwidth]{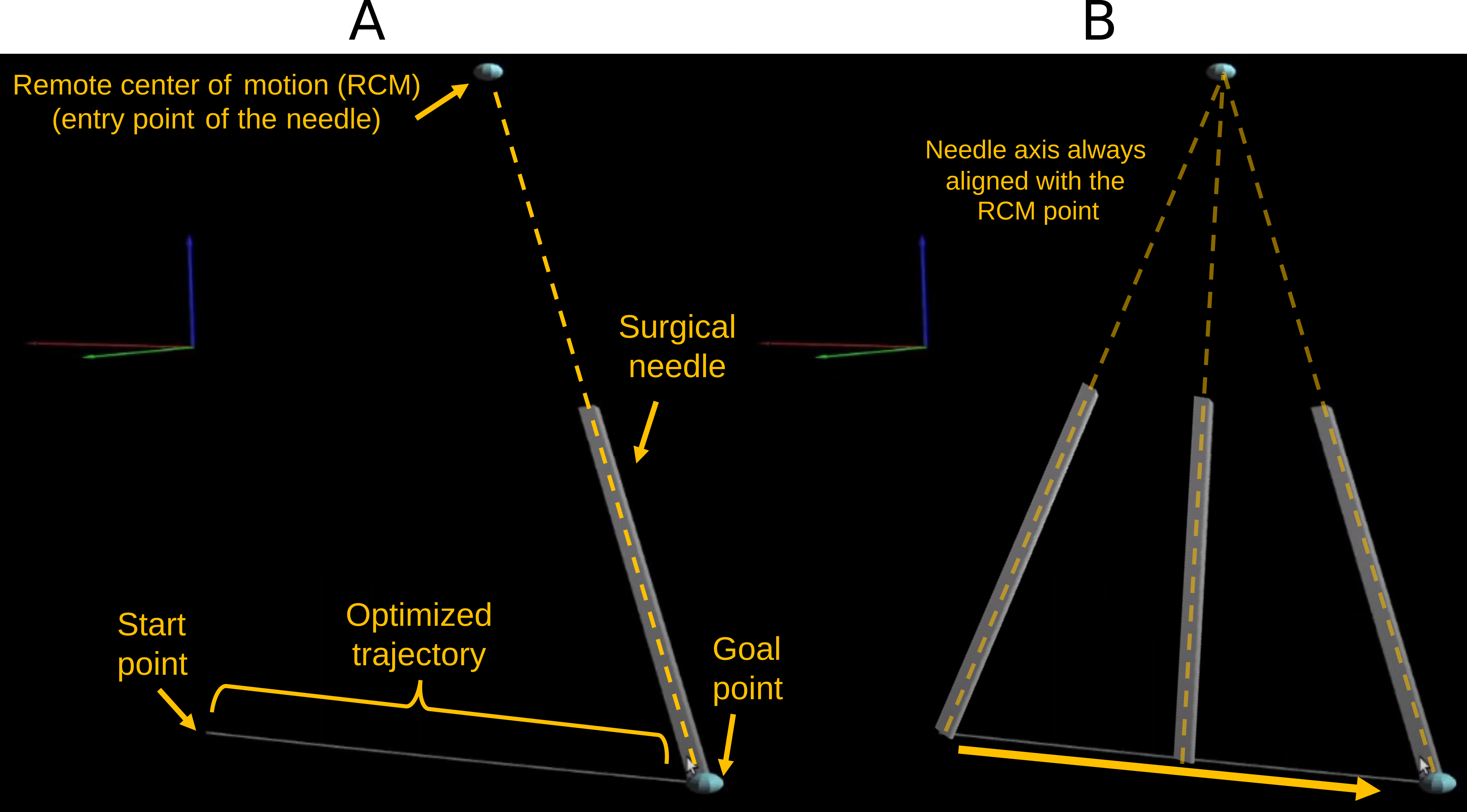}
        \caption{\small  (A) Illustration of trajectory optimzation; the RCM constraint must be satisfied at all times while reaching the desired goal in a smooth manner. (B) The optimized trajectory is shown.}
        \label{fig:ddp}
\end{figure}

\subsection{Optimal Control Formulation} \label{optimal_control}

Once a desired goal waypoint, $p_{desired}$,  is determined it is used by an optimal control framework to generate trajectory to the goal (Fig. \ref{fig:ddp}). Formally, we seek to generate an end-effector trajectory $x([t_0,t_f])$ over some time-interval $[t_0,t_f]$:
\begin{align}
    &  \arg\!\min_{\!\!\!\!\!\!\!\! x(\cdot),u(\cdot)} \int_{t_0}^{t_f} C(x(t), u(t)) dt, &&\, \text{:minimize cost } \label{eq:problem_cost} \\
    & \dot x(t) = f(x(t), u(t)), &&\, \text{:system dynamics} \label{eq:problem_dynamics}\\
    & (I-r_z(t) r_z(t)^T)(p_{rcm} - p(t)) = 0,  &&\, \text{:sclera constraint} \label{eq:problem_sclera}
\end{align}
where $C(x,u)$ is a given cost function e.g. ensuring smooth motion, and $r_z= Re_z$ and $e_z=(0,0,1)$ is the third basis vector (i.e. the surgical tool's longitudinal axis in the end-effector frame). We model the tool-tip as a fully-actuated rigid body with dynamics defined as

\begin{align}
    & \dot{g}(t) = g(t)\widehat{V(t)} \\
    & \dot{V}(t) = F(g(t), V(t), u(t)),
\end{align}
where $F$ encodes the effects of control forces, $u \in \mathbb{R}^6$, acting on the rigid body. The body velocity is defined as $\widehat{V} = g^{-1} \dot{g}$ where the "hat" operator in $\widehat{V}$ is defined as
\begin{align}
\widehat{V} =  \begin{bmatrix}
\widehat{w} & v \\
0_{1 \cross 3} & 0 
\end{bmatrix} , \, 
\widehat{\omega} =
\begin{bmatrix}
0 & -\omega_3 & \omega_2 \\
\omega_3 & 0 & -\omega_1 \\
-\omega_2 & \omega_1 & 0
\end{bmatrix}
\end{align}
and the components of $F$ are given as
\begin{align}
\dot{v_1} &= u_1 / m \\
\dot{v_2} &= u_2 / m \\
\dot{v_3} &= u_3 / m \\
\dot{\omega_1} &= \frac{(J_2 - J_3) \omega_2 \omega_3 + u_4}{J_1}\\
\dot{\omega_2} &= \frac{(J_3 - J_1)\omega_1 \omega_3 + u_5}{J_2}\\
\dot{\omega_3} &= \frac{(J_1 - J_2)\omega_1 \omega_2 + u_6}{J_3}
\end{align}
where $u_{0:2}$ are the linear and $u_{3:5}$ are the angular acceleration inputs along the body-fixed axes, $m$ is the mass of the system, and $J_{i}$ is the principal moment of inertia along each component. To solve this optimal control problem reliably in real-time we re-formulate it to include the constraint ~\eqref{eq:problem_sclera} as a least-square penalty, according to:

\begin{align}\label{opt_ctrl}
\begin{split}
C=&\frac{1}{2} \bigr[ {\| p_f - p(t_f)\|_{P_{p_f}}^2 + \|log(R_f^T R_{t_f})\|_{P_{R_f}}^2}\bigr] \\
&+\int_{t_{0}}^{t_{f}} \frac{1}{2} \|u(t)\|^2_{R} \\
& + w_s \cdot \| (I-r_z(t) r_z(t)^T)(p_{rcm} - p(t))\|^2 \mathrm{d}t,
\end{split}
\end{align}
where $p_f = p_{desired}$ and $R_f = R_{desired}$. Specifically, $p_{desired}$ is determined following the formulation given in Section \ref{calibration}, $R_{desired}$ is the orientation at $p_{desired}$ such that surgical tool's longitudinal axis is aligned with the RCM point (determined via simple trignometric relationships). In general, the overall cost aims to minimize the error in reaching the goal (encoded using $P_{p_f}, P_{R_f}\geq 0$ gain matrices), control effort (using $R>0$ gain matrix) and penalize deviation from the sclera point (using weight $w_s$). The optimal control problem is solved numerically using differential dynamic programming (DDP) based on a discrete-time quantization of the robot motion using some fixed time-step $dt$, over which the discrete controls are assumed to be constant \cite{geometric_opt_ctrl}. The resulting optimization generates a discrete sequence of states and controls, $x_{0:N_{traj}} \triangleq  \{ x_0, \cdots , x_{N_{traj}} \} $ and $u_{0:N_{traj}-1} \triangleq  \{ u_0, \cdots , u_{N_{traj}-1} \}$. The optimized trajectory is then used in a low-level controller to track the trajectory.


\subsection{Robot Control} \label{robot_control}

In this section, we describe how the optimal trajectory generated by the optimal control framework is tracked. Our robotic system is a 5-degree-of-freedom (DoF) robot, consisting of three translational bases and two rotational joints. Using the product of exponentials formula, the forward kinematics of the robot is given as the following \cite{mls_book}:

\begin{align}
g(q) = e^{\hat{\xi}_1 q_1 ... \hat{\xi}_5 q_5} g(0)
\end{align}
where $q_i$, $i = 1, ..., 5$ are the joint angles of the robot, $\xi_i$ is the twist representation associated with the rigid-body transformation of the $i$th joint, and $g(0)$ is the relative transformation between the robot base and end-effector frame when $q = 0_{5 \cross 1}$. The robot joint velocities can be related to the end-effector velocity using the body manipulator jacobian:

\begin{align}
V = J_{r}(q) \dot{q} \label{eq:body_jacobian}
\end{align}
where $J_{r}$ is the body manipulator jacobian and $\dot{q}$ are the joint angle velocities. The body manipulator jacobian is given as the following:

\begin{align}
J_r(q) &= [ \xi_{1}^{\dagger}  \cdots \xi_{5}^{\dagger}] \\
\xi_{i}^{\dagger} &= Ad^{-1}_{(e^{\hat{\xi}_i q_i} \cdots e^{\hat{\xi}_{5} q_{5}} g(0))}\xi_{i}
\end{align}

The optimized trajectory, $x_{0:N_{traj}}$, is tracked in the robot task space. Specifically, among the list of states in $x_{0:N_{traj}}$, the desired state or state-to-track is determined by choosing the index with the closest pose to the current end-effector pose with some look-ahead offset. The end-effector velocity associated with the chosen state is converted to robot joint velocities via Eq. \ref{eq:body_jacobian}, which the low-level robot controller tracks via PID control. As the end-effector traverses through the optimal trajectory, the index of state-to-track is updated accordingly.

\begin{figure}[h]
        \centering
        \includegraphics[width = \columnwidth]{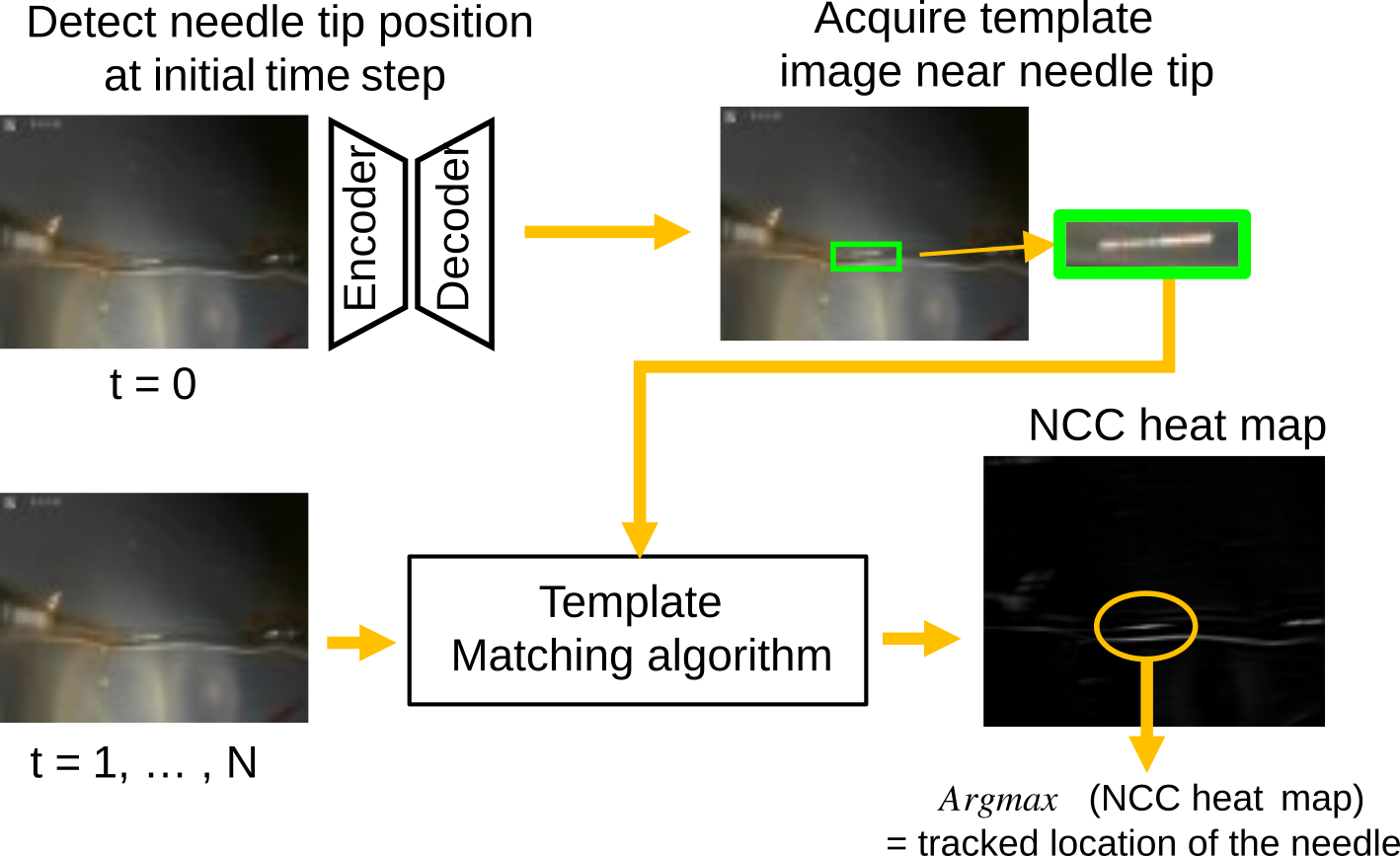}
        \caption{\small  Illustration of the contact detection pipeline. The needle appearance is compared to its appearance in the future time steps as it is lowered toward the target vein. $t = 0$ is the time step which the needle lowering step begins. Upon close inspection, the NCC heat map resembles the input image. The brightest spot in the heat map is the tracked location of the needle based on its similiarity to the template image.}
        \label{fig:contact_detection_pipeline}
\end{figure}

\begin{figure}[h]
        \centering
        \includegraphics[width = \columnwidth]{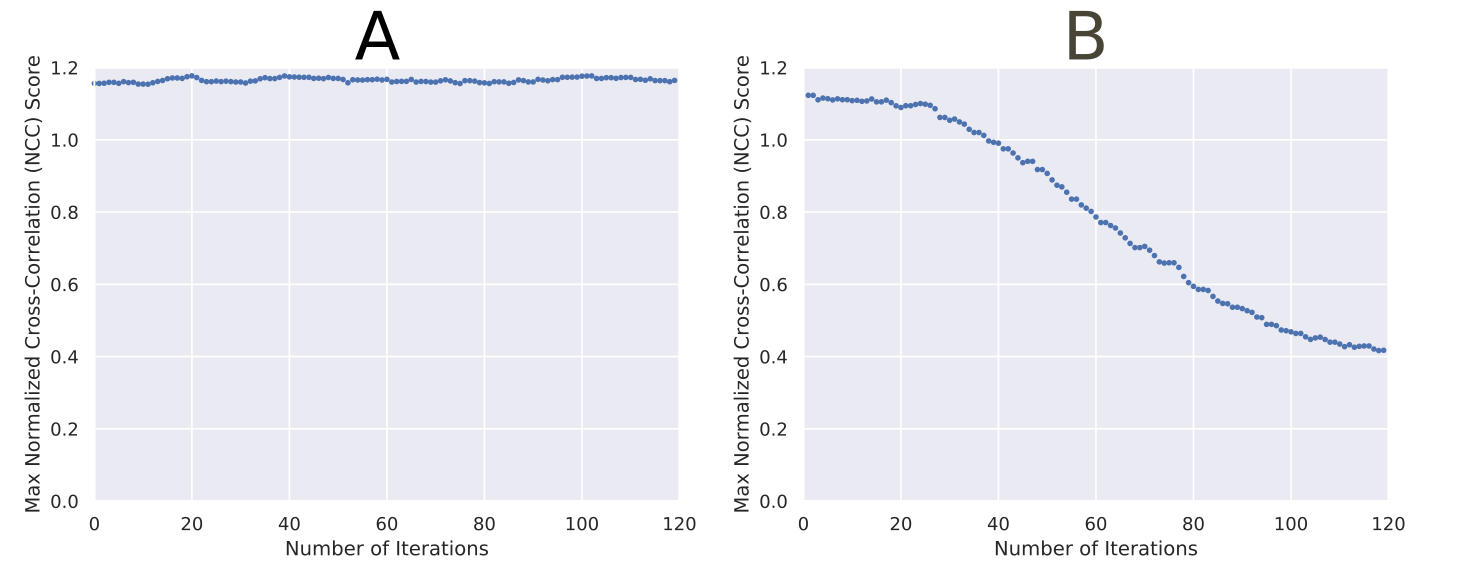}
        \caption{\small  Comparison of maximum NCC scores as the needle is lowered. In (A), no contact between the needle and the target vein is observed therefore the max NCC score remains relatively consistent. In (B), contact is observed, which causes a change in the needle's appearance, therefore causing the max NCC score to drop. Such decreasing trend used is as a cue to detect contact between the needle and the target vein.}
        \label{fig:contact_detection_no_contact_vs_contact_graph}
\end{figure}

\subsection{Detecting Contact between the Needle and the Vein} \label{contact_detection_section}

Recall that once the needle-tip is aligned with the clicked goal in the image view, the subsequent action is to lower the needle towards the target vein. Once the needle-tip contacts the target vein, the needle deflects, which causes a visible change in its appearance\footnote{There are other cues observed when the needle tip contacts the vein, such as the local dimpling of the tissue. In the present work, we focused on the visual change of the needle}. This cue is detected using a classic template matching algorithm \cite{template_matching_book}, as illustrated in Fig. \ref{fig:contact_detection_pipeline}. At a high level, the template matching algorithm is a method for locating small parts of an image which match the given template image, based on a chosen metric.

As the needle is lowered, the template matching algorithm continues to track the needle, but with a lower match score as the needle deflects and changes appearance as shown in Fig. \ref{fig:contact_detection_no_contact_vs_contact_graph}. To be clear, the match score drops because the needle no longer appears similar to the given template image. If the match score decreases by a significant threshold (i.e. the needle significantly changes its appearance), a contact event is detected and the robot is stopped. Specifically, the template matching algorithm generates a heat map where the index of the highest score is the tracked location of the needle. For the similarity metric, we use the normalized cross-correlation (NCC) metric, which is given by,

\makeatletter
    \def\tagform@#1{\maketag@@@{\normalsize(#1)\@@italiccorr}}
\makeatother

\footnotesize
\begin{equation}
    NCC_t[i,j] = \frac{\Sigma_m \Sigma_n (a[m,n] - \bar{a}) (b[m,n] - \bar{b})}
    {\sqrt{\Sigma_m \Sigma_n (a[m,n] - \bar{a})^2} \sqrt{\Sigma_m \Sigma_n (b[m,n] - \bar{b})^2}}
\end{equation}
\normalsize
where $a$ and $\bar{a}$ denote the template image and its mean, $b$ denotes the ROI image from a subsequent time step cropped to the same size as the template image and $\bar{b}$ denotes its mean, $m$ and $n$ are the index along width and height of the comparing images, and $i$ and $j$ are the index along the width and height of the output NCC heat map. Intuitively, the NCC metric is a normalized dot product between the template image and the comparing image. Considering each image as vectors, the dot product between them are high when the two vectors are similar (i.e. the images appear similar) and low when the two vectors are different (i.e. the images appear different).

While the needle is lowered toward the target vein, the NCC score is calculated until contact is detected. The percent change in maximum NCC score compared to its initial score is given by:

\begin{equation}
    f(NCC_{t_0}, NCC_t) = \frac{max(NCC_{t_0}) - max(NCC_{t})}{max(NCC_{t_0})}
\end{equation}
\normalsize

where $NCC_{t_0}$ is the NCC heat map at the start of the needle lowering motion and $NCC_t$ is the NCC heat map calculated in subsequent time steps. If the percent change of maximum NCC score is greater than a chosen threshold gain, then the algorithm detects a contact event. In other words,

\begin{equation}
    isContact = 
\begin{cases}
    True ,& \text{if }  f(NCC_{t_0}, NCC_t) \geq \gamma \\
    False,              & \text{otherwise}
\end{cases}
\end{equation}

The threshold gain $\gamma$ was chosen by observing 14 representative vein-contact events as examples. Specifically, we manually chose a value such that contact events could be detected with a small delay in these examples.

\subsection{Cannulation and Detecting Vein Puncture} \label{puncture_detection}
Once the needle makes slight contact with the vein, the needle is cannulated along its axis in a smooth motion. In practice, once a significant travel distance is achieved, the needle punctures the vein wall. The puncture event leads to a sudden release of potential energy which accelerates the needle into the lumen, which induces the risk for double-puncturing the vein. It is therefore important to stop advancing the needle as soon as the puncture is detected. 

To detect puncture, we collected many examples of puncture events to design a logic-based way of detecting punctures. However, this task proved to be challenging because there were too many variations in terms of visual appearance and speed at which puncture events occurred. Therefore, we chose a data-driven approach of training a recurrent CNN to detect puncture events from images. Specifically, we used a Resnet-18 \cite{resnet} as an encoder network which used a microscope image ($224 \times 224 \times 3$) as input and generated a binary output denoting the probability of puncture. To improve the network's performance, we added a decoder network on top of the encoded features to predict the input image, mimicking an auto-encoder network. Such use of auxiliary loss added an additional gradient signal that improved the network's performance on the main task of detecting punctures. At the time of inference, however, the decoder was ignored and thus it did not add any additional computational overhead during inference. We express the loss function as 

\begin{equation}
\begin{split}
    L((y, I), (\hat{y}, \hat{I})) =  & -\frac{1}{N} \sum_{n=1}^{N} \{ y_i \cdot log(\hat{y_i}) \\
    & + (1-y_i) \cdot log(1-\hat{y_i}) + (I - \hat{I})^2 \}
    \end{split}
\end{equation}
where $y$ is the binary indicator for the true class, $\hat{y}$ is the predicted probability of puncture, $I$ is the input image, and $\hat{I}$ is the image predicted by the decoder part of the network. We define $p_c = \hat{y}$ and $p_{vp} = 1 - \hat{y}$ as the probability of contact and venipuncture respectively. Note that the puncture detection module is activated after detecting a needle-vein contact event. Thus, $p_c$ is not used to detect needle-vein contact events. It simply denotes the probability of the current surgical state being the surgical state preceding the event of puncture.

During training, a time horizon of 15 time steps was used, which is equivalent to three seconds of video. Approximately 200 puncture events were used for training, and 50 for testing. To improve the network's accuracy, some additional changes were necessary. We sub-sampled input video from 30Hz to 7Hz so that the puncture event would appear more obvious between frames. This was due to some vein punctures occurring very subtly and slowly across several frames. Furthermore, the network was trained to detect punctures 1 or 2 frames after the actual event of puncture. This prevented the network from triggering early, which could cause the robot to stop before completion of the actual puncture event.

\subsection{Needle tip detection} \label{tool_tip_detection}
The objective is to detect the needle tip position in image coordinates, given an input image dimension of $640 \cross 480 \cross 3$. Similar to the puncture detection network, we use
Resnet-18 as the backbone in an encoder-decoder network to predict the coordinates of the tool-tip. The output dimension of the network is a single channel image with identical width and height as the input ($640 \cross 480 \cross 1)$. The loss function is given as

\begin{align}
    L((u, \hat{u}), (v, \hat{v})) =  & -\frac{1}{N} \sum_{n=1}^{N} u_i \cdot log(\hat{u_i}) + v_i \cdot log(\hat{v_i})
\end{align}

where $u$ is the true class along the image width and $\hat{u}$ is the prediction ($u, \hat{u} \in \mathbb{Z} \colon 0 \leq u, \hat{u} < 640$), and $v$ is the true class along the height of the image and $\hat{v}$ is the prediction  ($v, \hat{v} \in \mathbb{Z} \colon 0 \leq v, \hat{v} < 480$). Approximately 1200 images were manually labelled for training an 300 images were used for testing. For improved robustness, data augmentation was used. In particular, we performed random crop and resizing of the image near the tool-tip, which simulated the effect of the tool being out-of-focus. This augmentation enabled detection of the tool-tip even if the needle-tip was out-of-focus, which was frequently encountered during the experiments. Random rotations were also necessary so that the needle could be detected in various orientations, among other common techniques such as pixel dropout and hue jitter. 

\begin{figure}[h]
        \centering
        \includegraphics[width = \columnwidth]{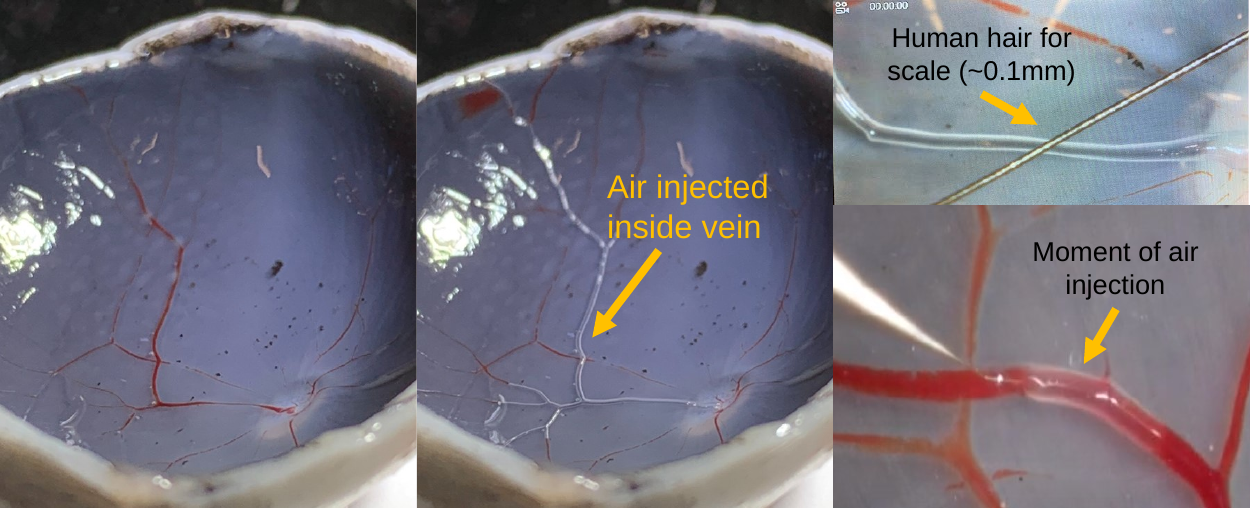}
        \caption{\small  (Left) Pig eye after anterior dissection and vitreous removal. (Middle) Pig eye after air injection into the retinal vein to restore vascular pressure. (Right, top) Human hair of approximately 0.1mm in diameter is compared against the vein. (Right, bottom) Moment of air injection is captured.}
        \label{fig:pig_eye}
\end{figure}

\begin{figure*}[h]
        \centering
        \includegraphics[width=0.98\textwidth]{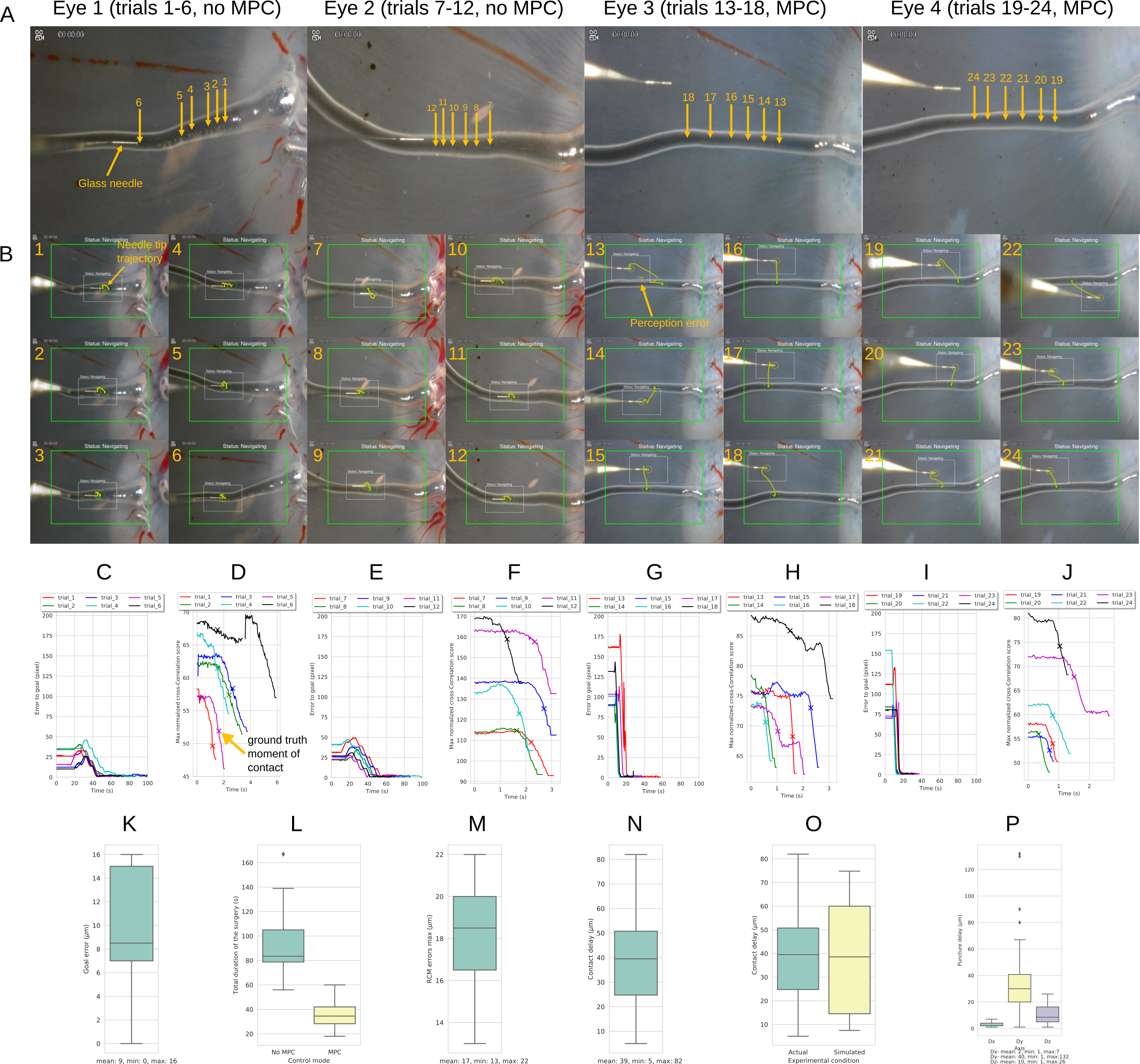}
        \caption{\footnotesize  (A) Results of the 24 trials of autonomous RVC are shown; 6 trials were performed on 4 different pig eyes. First 12 trials were performed without MPC and the remaining trials used MPC (B) The trajectory of all trials are shown in image coordinates. (C, E, G, I) The pixel error to the clicked goal is shown through time. (D, F, H, J) The maximum NCC score is shown as the needle contacts the vein. The moment at which the human could detect a contact event is labelled, which we regard as the ground-truth moment of contact. The last plotted points denote the point at which our system detected a contact event and stopped the robot. (K) The final error to the clicked goal is shown. (L) Total duration of surgery is compared between no-MPC and MPC mode. (M) Maximum RCM errors for all trials are reported. (N) Needle-vein contact delay is reported, (O) Contact delay is compared between the actual experiment and in the simulated experiment. (P) Extra distance travelled by the needle-tip due to puncture delay is reported.} 
        \label{fig:main_results}
\end{figure*}

\section{Experimental Setup}

The experimental setup consists of the Steady Hand Eye Robot (SHER) \cite{ER_2_0}, a glass micropipette attached at the end-effector, and a microscope recording the top-down view of the surgery. SHER is a surgical robot developed specifically for retinal surgery applications. For the glass needle, we used ICSI Origio micropipettes (Ballerup, Denmark), with an inner diameter of 5.0 - 5.5$\mu m$ and a 45 degree bent elbow. Its outer diameter is unspecified, though by visual inspection it appears to be $\sim$10 - 15$\mu m$. To increase the visibility of the glass needle, it was coated with a thin layer of silver which was similarly done in human RVC trials \cite{first_human_retinal_surgery}. A syringe filled with air was connected to the needle and injected after each needle insertion to confirm successful injections. The porcine eye cups were prepared by cutting along its equator using a surgical scissor and removing the internal vitreous. To enable multiple cannulations on a single eye, the veins were pressurized by manually cannulating the vein and injecting air using a syringe, as shown in Fig. \ref{fig:pig_eye}.

\section{Experiments}

To test the efficacy of our system, we conducted a total of 24 needle insertion trials on 4 different pig eyes. For each eye, 6 trials were performed. The trials for each eye were performed back-to-back, without stopping between trials, to evaluate the system's unbiased performance. The following describes the procedure for each trial: after pig eye preparation, the human operator navigated the surgical tool near the target vein, while ensuring that the needle was roughly placed above the retinal surface. Then, the operator initialized a virtual RCM point 2cm along the axis of the tool-tip via a mouse-click in the GUI to simulate the entry point of the eye. Note that in practice, the RCM point is recorded at the moment of entry when the needle is inserted through the sclera. The operator then specified a 2-d target insertion point in the microscope image via a mouse click. The operator then clicked "start", after which the autonomous procedure commenced. All the steps including navigation to the target vein, placement of the needle on the vein, needle insertion, and stopping the robot after detecting puncture were performed autonomously without further intervention from the operator.

As a mode of comparison, we compared our autonomous vein cannulation system to robot-assisted and free-hand trials performed by a human. During robot-assisted mode, the user cooperatively controlled the robot by yielding the tool attached at the end-effector while modulating the gain of the motion using a foot pedal. In free-hand mode, no robotic assistance was provided, and the hand was anchored on a platform for stability.

\begin{figure*}[h]
        \centering
        \includegraphics[width = \textwidth]{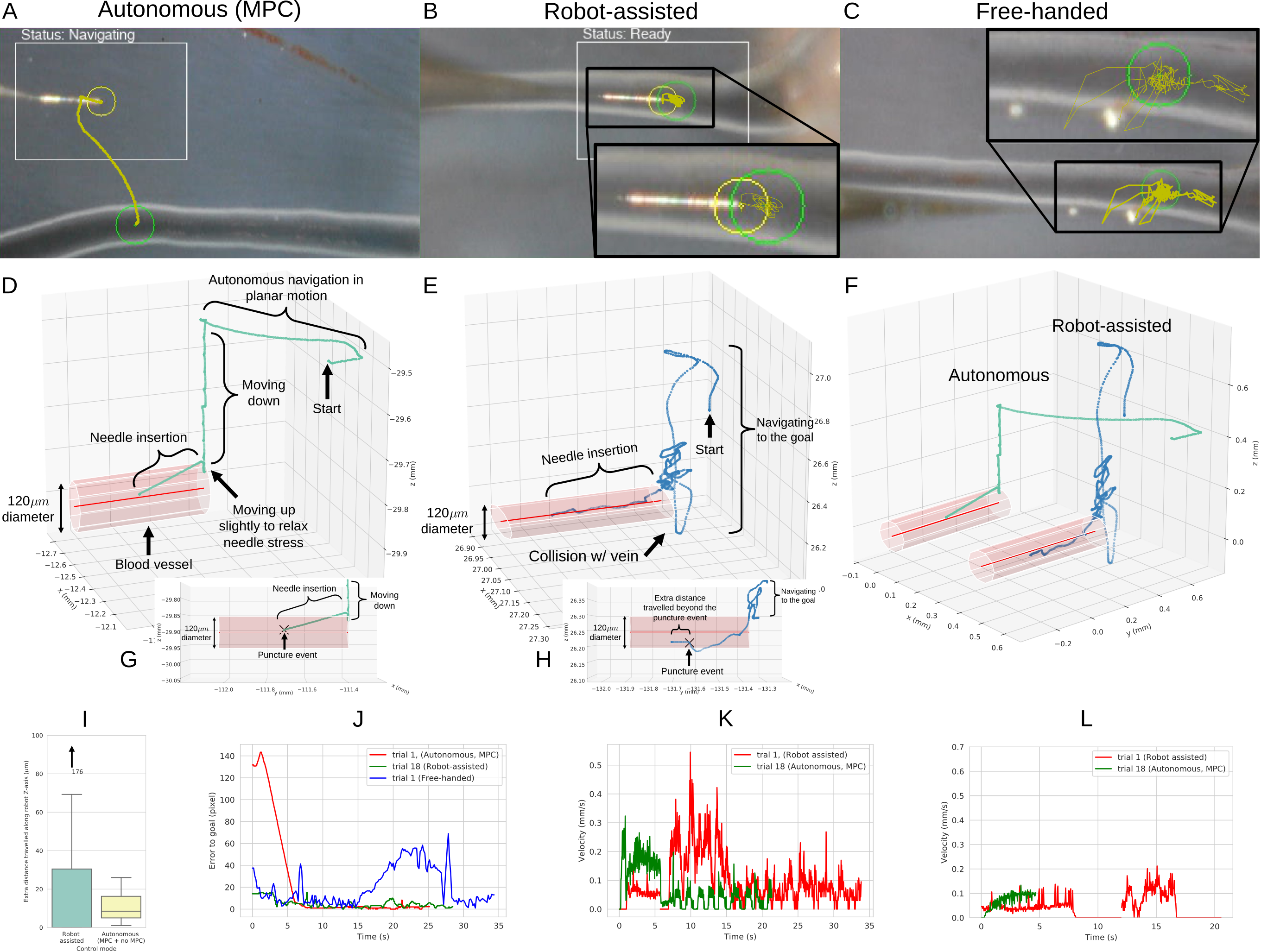}
        \caption{\small Navigation trajectory is shown for (A) autonomous, (B) robot-assisted, and (C) free-hand modes. (D, E) The trajectories of the autonomous mode and the free-hand mode is shown in 3D. (F) Their trajectories are compared side-by-side. (G, H) A close-up of the insertion trajectory is shown from a side view (I) Extra distance travelled along the Z-axis after the puncture is compared for robot-assisted and autonomous trials. (J) The error to the final clicked goal is compared between robot-assisted and autonomous mode. (K) The velocity of the needle navigation trajectory is compared between robot-assisted and autonomous mode. (L) Similar comparison is made during needle insertion}
        \label{fig:comparison_results}
\end{figure*}

\section{Results and Discussion}
\subsection{Navigation Performance}
The navigation results for all 24 trials are shown in Fig. \ref{fig:main_results}. Fig. \ref{fig:main_results}A shows the 6 needle insertion locations for each eye, which are also the target needle insertion sites which the user specified via mouse-clicks during the experiment.  Fig. \ref{fig:main_results}B shows the trajectory followed by the needle in image coordinates. Fig. \ref{fig:main_results} C, E, G, I show the pixel error between the tool-tip and the clicked goal throughout the needle navigation procedure.

Note that trials 1-12 did not use model predictive control (MPC) and trials 13-24 used MPC. The first 12 trials did not use MPC as our primary objective was to test the system in the initial trials. Thus, in the first 12 trials, the tool was navigated more slowly and the needle-tip was initialized closer to the goal, as shown in Fig. \ref{fig:main_results}B, C, and E. Fig. \ref{fig:main_results}K shows the needle-tip placement accuracy on the clicked goals. For all trials, the mean needle-tip placement error was 9$\mu$m and the maximum error was 16$\mu$m. Such level of accuracy meets the 20 - 30 $\mu$m accuracy requirement for needle-tip placement accuracy during retinal surgery. Note that the metric errors were acquired by converting the observed pixel error using a conversion factor of 136.33pixels/mm. 

In Fig. \ref{fig:main_results}B, it can be observed that the trajectories are not straight-line paths from the initial position to the goal. This is due to the jacobian (i.e. the calibration matrix used during
visual servoing) being estimated in real-time. Specifically, the
jacobian was initialized to an identity matrix (i.e. initialized to an
arbitrary non-singular matrix), but eventually it converged to
the true jacobian after observing a few robot motions. Thus,
the trajectory was not initially directed towards the goal
but eventually the goal was reached as the jacobian converged. In practice, the jacobian converged
within 2-3 seconds. We also found that using such a real-time calibration approach (i.e. IBVS) facilitated repeated experimental trials without interruption. When the magnification or the position of the microscope was accidentally perturbed or changed for fine adjustments, recalibration was not necessary.

We also report the total duration of all trials Fig. \ref{fig:main_results}L, which was measured starting from the point when the operator clicked "start" until the stopping of the robot after detecting venipuncture. In no-MPC mode (trials 1-12), we were mostly interested in testing the system thus the overall procedure was performed slowly. However, in MPC mode (trials 13-24), the entire procedure could completed within 35 seconds on average, and at maximum under 1 minute. The RCM errors were kept low, as shown in Fig. \ref{fig:main_results}M. Specifically, for all trials, the RCM error never exceeded more than 22$\mu m$, which meets the safety requirements for retinal surgery. Note that the RCM error was calculated using Eq. \ref{eq:problem_sclera} using precise readings from the robot kinematics. The calculated RCM errors were also displayed in the GUI (Fig. \ref{fig:gui}) in real-time so that the user could verify that the constraint was being satisfied during the surgery.

\subsection{Contact Detection Performance} \label{contact_detection_performance}Fig. \ref{fig:main_results}D, F, H, and J illustrate the decreasing NCC scores during contact detection. Recall that when the needle is lowered and the tip contacts the vein, the needle's appearance changes, which causes the template matching algorithm to track the needle with a lower NCC match score. The decreasing trend shown in these plots reflects the decreasing match score as the needle contacts the vein. To verify that the contact event was detected with an appropriate timing, we reviewed the recorded videos after the experiment and labelled the exact moment at which the human could detect a contact event\footnote{Note that the human labels are not reflective of human performance, since in a real-life scenario, humans would respond to the contact event after some delay (e.g. after a contact event is perceived, a button to stop the robot may be pressed after 0.25 second delay or within some human reaction time).}. We regard the human labels as the ground-truth moments of contact and they are labelled with a marker in the mentioned figures. Additionally, the last plotted point in each trial is the moment at which our system detected a contact event and stopped the robot.

Our results show that in all trials, the human could detect contact events before the system could detect them, although they are very close. As a result of temporal delay in detecting contact, we quantitatively show the extra distance travelled by the needle-tip beyond the ground-truth point of contact in Fig. \ref{fig:main_results}N.  Across all trials, the observed mean extra distance travelled was 39 $\mu$m and the maximum was 82 $\mu$m. Due to the extra distance travelled, we did not observe any injury to the retinal veins as the needle motion was not axial along the needle's sharp tip (since the needle is lowered along the robot's spatial Z-axis, without any XY motion), therefore the vein was only pushed. Furthermore, the needle-tip is very flexible and can easily pivot about its elbow. Therefore, the extra distance travelled by the needle does not reflect the amount of tissue deformation, but rather a combination of needle deflection and tissue deformation. 

To tune the threshold gain of the contact detection system (see Section \ref{contact_detection_section}), we considered 14 contact examples. Fig. \ref{fig:main_results}O compares the expected contact delay based on these 14 examples and the actual contact delay observed in the 24 trials. The figure shows that larger delays were observed during the experiment. Therefore, more contact examples should have been used to tune the contact threshold gain.

\subsection{Puncture Detection Performance}
Fig. \ref{fig:main_results}P shows the venipuncture detection performance in terms of extra distance travelled by the needle due to the temporal delay in detecting puncture. In Fig. \ref{fig:main_results}P, Dx, Dy, and Dz refer to the extra distance travelled along the robot's spatial X, Y, and Z axes respectively beyond the puncture point. To measure this, we reviewed the video recordings and marked the exact moment at which puncture occurred, which we regarded as the ground-truth moment of puncture. Then, we compared the tool-tip position at the moment of ground-truth puncture against the tool-tip position when the robot was stopped after detecting puncture to compute Dx, Dy, and Dz.

For ease of understanding, the directions Dx and Dy can be considered to form a plane that is parallel to the surface of the target vein. Thus, of most interest is the extra distance travelled along the Z-axis of the robot (Dz) or towards the second wall of the vein\footnote{The assumption that Dx and Dy are parallel to the surface of the target vein implies that the vein is flat w.r.t the robot spatial frame. In practice during RVC, the needle-tip is cannulated downhill towards the basin of the eye. This scenario is safer since more travel distance along the Z-axis would be required to reach the second wall of the vein.} Ideally, the Dz should be as low as possible to avoid the risk of double-puncturing the vein. As shown in Fig. \ref{fig:main_results}P, the mean extra distance travelled along the Dz direction was 11$\mu$m and the maximum was 26$\mu$m. Considering that the scale of the targeted veins range from 80 - 120$\mu m$ in diameter, a maximum puncture delay of 26$\mu m$ may be considered safe. Furthermore, after each trial, we injected air using a syringe to verify that double-puncture did not occur.

We note that there are a few sources of delay in detecting puncture. One contributing factor is our design choice since we sub-sampled the input video into the network from 30Hz to 7Hz to make the puncture events more visibly obvious between frames. Also, the puncture event labels were deliberately delayed to avoid early detections. Another source of delay may be the novel punctures observed which are not contained in the distribution of the training dataset. In practice, we observed that punctures that occurred very subtly and slowly across several frames were detected with greater delay. Veins that "popped" more dramatically and were more visibly obvious to the human eye were detected almost immediately, which is expected.

\subsection{Comparison between autonomous, robot-assisted, and free-hand trials}

Fig. \ref{fig:comparison_results} illustrates comparisons between representative examples of autonomous mode and a human performing in robot-assisted and free-hand mode. 32 robot-assisted trials and 8 free-hand trials were performed. Fig. \ref{fig:comparison_results} A, B, C shows their trajectories plotted in image coordinates. The trajectory during autonomous mode is concise and efficient, while robot-assisted and free-hand trajectories show hesitation and instability in reaching the goal. This instability can also be visualized in \ref{fig:comparison_results}J, which shows the pixel error between the tool-tip and the goal fluctuating over time. Even robot-assisted mode posed challenges because as the needle approached the goal, the operator had to slow-down gradually by carefully reducing the pedal gain and the force exerted on the robot end-effector. If the pedal gain or the force exerted on the end-effector was not gradually released at the correct moment, the needle overshot the goal and multiple attempts were necessary. Vein cannulation in free-hand mode was limited by hand tremor, where it was challenging to keep the needle steady on the pixel goal or insert the needle perfectly along its axis.

Fig. \ref{fig:comparison_results}D, E, and F compares the autonomous and robot-assisted trajectory in 3-d. Fig. \ref{fig:comparison_results}G and H shows a close-up view of the needle insertion step, which also shows the ground-truth point of puncture. The moment at which the robot was stopped after detecting puncture is the last plotted point in the trajectory. For autonomous mode (Fig. \ref{fig:comparison_results}G), these two points nearly coincide as the robot is immediately stopped after detecting puncture. However, in robot-assisted mode  (Fig. \ref{fig:comparison_results}H) the human operator inevitably travels extra distance due to human reaction time in stopping the robot. The comparison between extra distance travelled along the Z-axis of the robot (i.e. toward the other wall of the vein) beyond the ground-truth puncture point is compared in Fig. \ref{fig:comparison_results}I. This figure shows that the extra distance travelled during autonomous mode is less compared to robot-assisted mode. Specifically, the mean extra distance travelled during autonomous mode 11 $\mu$m, while 20 $\mu$m was observed in robot-assisted mode. Furthermore, during robot-assisted mode, the human sometimes collided with the target vein during navigation due to lack of depth perception. However, these events were rare and did not cause visible damage to the target vein. We also compared the robot velocity profile during navigation and needle insertion steps in Fig. \ref{fig:comparison_results}K and L, which showed that during autonomous mode, the robot was operated with a significantly lower velocity profile and with predictable consistency. 

\section{Conclusion}
In this work, we demonstrate an autonomous system for RVC. Our system requires minimal setup and guidance, requiring monocular images as input and goal specified by the user via clicking. We showed that our system could accurately navigate the surgical needle onto the target vein within required margins of safety, successfully perform needle insertion, and detect a venipuncture event in a timely manner to avoid double-puncturing the vein. We demonstrated the consistency of our system and its ability to generalize  across various pig eye anatomy. 


In this work, we did not consider the patient's eye motion. During the surgery, the patient has motion related to breathing, which causes cyclic oscillation of the retinal tissue. This may not be a critical limitation since in recent robot-assisted human trials, the needle was held static inside the vein without motion compensation \cite{first_human_retinal_surgery}. The surgeon, however, supervised the infusion procedure and only when the patient moved significantly, was the robot manually removed from vein to avoid damage to the retina.

A future point of improvement is reducing the complexity of our system. In our pipeline, we had to manually enumerate the main events of RVC and encode them into surgical states and transitions between states. This required manual hand-crafted work. A more general approach may be to use a temporal CNN that consumes images as input and outputs robot actions. However, this would likely require an infeasible amount of data and its ability to operate safely in an unseen environment would be questionable. Recent work has shown promise in this direction by training a CNN to perform surgical tool navigation in retinal surgery, though without performing needle insertion \cite{autonomous_navigation_retina}, \cite{eye_surgery_imitation_learning}, \cite{peiyao_eye_surgery}, \cite{peiyao_pig_eye_paper}.  

Another consideration is the extension of this work to a variety of surgical needles. In our work, we used small-diameter glass needles, which was easily pliable. This property was beneficial in designing a sensitive contact detection system. Less pliable needles such as metal needles also demonstrate similar contact cues. However, other visual cues may have to be considered, such as tissue deformation.

Future work will consider transitioning to a more realistic eye model, such as closed pig eyes and live animals.

\bibliography{bib}  



\end{document}